%% file: main.tex
\pdfoutput=1

\documentclass[11pt]{article}

\usepackage[]{ACL2023}
\usepackage{makecell}
\usepackage{caption}
\usepackage{multirow}
\usepackage{enumitem}
\usepackage{graphicx}
\usepackage{times}
\usepackage{latexsym}
\usepackage{booktabs}
\usepackage{xspace}
\usepackage{subcaption}
\PassOptionsToPackage{table}{xcolor}
\usepackage[table]{xcolor}
\usepackage[T1]{fontenc}

\usepackage[utf8]{inputenc}

\usepackage{microtype}

\usepackage{inconsolata}

\usepackage{times}
\usepackage{latexsym}
\usepackage{url}            %
\usepackage{wrapfig}
\usepackage{booktabs}       %

\usepackage{caption}
\usepackage{placeins}
\usepackage{amssymb}%
\usepackage{pifont}%
\usepackage{floatrow}
\usepackage{fancyvrb}
\usepackage{graphicx}
\usepackage{amsmath}
\usepackage{amsfonts}       %
\usepackage{nicefrac}       %
\usepackage{microtype}      %
\usepackage{xcolor}         %
\usepackage{xspace}         %
\usepackage{multirow}
\usepackage{array}
\usepackage{siunitx}
\usepackage{booktabs}
\usepackage{colortbl}
\usepackage{color}
\usepackage{soul}
\usepackage{tcolorbox}
\usepackage{hyperref}
\usepackage{highlight}

\def\Snospace~{\S{}}

\newcommand\extrafootertext[1]{%
    \bgroup
    \renewcommand\thefootnote{\fnsymbol{footnote}}%
    \renewcommand\thempfootnote{\fnsymbol{mpfootnote}}%
    \footnotetext[0]{#1}%
    \egroup
}

\usepackage[T1]{fontenc}
\definecolor{mygray}{RGB}{220, 220, 220}
\definecolor{lightgreen}{RGB}{223,255,219}
\definecolor{lightred}{RGB}{255,219,219}

\definecolor{pos}{RGB}{167, 199, 231}
\definecolor{neg}{RGB}{250, 160, 160}

\usepackage{soul}

\DeclareRobustCommand{\hlpos}[1]{{\sethlcolor{pos}\hl{#1}}}
\DeclareRobustCommand{\hlneg}[1]{{\sethlcolor{neg}\hl{#1}}}

\DeclareRobustCommand{\baseline}[1]{\cellcolor{gray!25}{#1}}

\usepackage[utf8]{inputenc}

\newcommand{\chatgpt}{\texttt{gpt-3.5-turbo-0613}\xspace}
\newcommand{\chatgptShort}{\texttt{GPT3.5}\xspace}
\newcommand{\mistral}{\texttt{Mistral}\xspace}
\newcommand{\gpto}{\texttt{GPT4o}\xspace}

\newcommand{\llamaShort}{\texttt{LLama2}\xspace}
\newcommand{\vicunaShort}{\texttt{Vicuna}\xspace}
\newcommand{\flanShort}{\texttt{Flan-T5}\xspace}

\input{commands}

\title{Characterizing LLM Abstention Behavior in Science QA with Context Perturbations}

\author{Bingbing Wen$^{1}$ ~~
Bill Howe$^{1}$ ~~
Lucy Lu Wang$^{1,2}$ \\
$^1$University of Washington, ~$^2$Allen Institute for AI \\
\texttt{\{bingbw, billhowe, lucylw\}@uw.edu}  \\
}

\begin{document}
\maketitle

\begin{abstract}
  \input{sections/abs_new}
\end{abstract}

\input{sections/intro_new}

\input{sections/related_work_new}
\input{sections/method_new}

\input{sections/exp_new}

\input{sections/discussion_new}

\input{sections/limitation_new}
\vspace{-2mm}
\section*{Acknowledgements}
\vspace{-2mm}
This research was supported in part by the UW iSchool Strategic Research Fund.


\bibliography{anthology,custom}
\bibliographystyle{acl_natbib}

\clearpage
\input{sections/appendix_new}



\end{document}

%% file: commands.tex
\newcommand\squad{SQuAD2\xspace}
\newcommand\qasper{QASPER\xspace}
\newcommand\pubmedqa{PubmedQA\xspace}
\newcommand\bioasq{BioASQ\xspace}

\newcolumntype{L}[1]{>{\raggedright\let\newline\\\arraybackslash\hspace{0pt}}p{#1}}
\newcolumntype{M}[1]{>{\raggedright\let\newline\\\arraybackslash\hspace{0pt}}m{#1}}

\definecolor{darkgreen}{rgb}{0.0, 0.4, 0.13}

%% file: sections/abs_new.tex
The correct model response in the face of uncertainty is to abstain from answering a question so as not to mislead the user.
In this work, we study the ability of LLMs to abstain from answering context-dependent science questions when provided insufficient or incorrect context.
We probe model sensitivity in several settings: removing gold context, replacing gold context with irrelevant context, and providing additional context beyond what is given. In experiments on four QA datasets with six LLMs, we show that performance varies greatly across models, 
across the type of context provided, and also by question type; in particular, many LLMs seem unable to abstain from answering boolean questions using standard QA prompts. 
Our analysis also highlights 
the unexpected impact of abstention performance on QA task accuracy. Counter-intuitively, in some settings, replacing gold context with irrelevant context or adding irrelevant context to gold context can improve abstention performance in a way that results in improvements in task performance. Our results imply that changes are needed in QA dataset design and evaluation to more effectively assess the correctness and downstream impacts of model abstention.\footnote{Code is publicly available at \url{https://github.com/bbwen/llm_scienceqa}.}

%% file: sections/intro_new.tex
\section{Introduction}

\begin{figure}[t!]
\centering
\minipage{\linewidth}
\includegraphics[width=\linewidth]{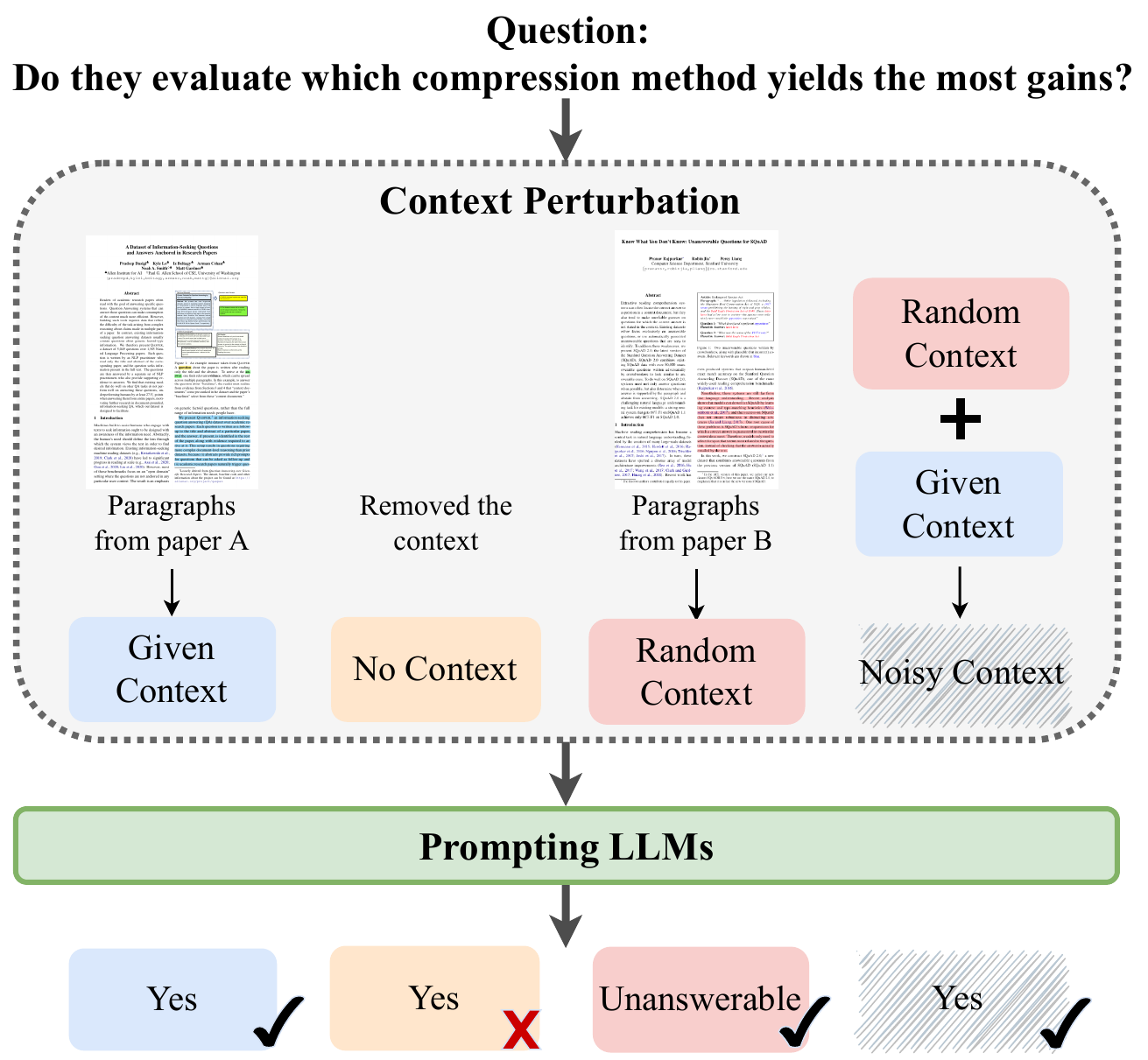}
\caption{Our framework to probe the context sensitivity of LLMs for science QA. We show an example from the QASPER dataset and the prediction results of \chatgptShort under different context perturbation settings. While the model fails to abstain when context is removed, it abstains appropriately when random context is provided.}\label{fig:framework}
\endminipage
\end{figure}


Question-answering (QA) in scientific settings 
is often defined as a context-dependent task, where models answer questions based on provided context or relevant context it identifies.
When the provided or retrieved context is itself unreliable or inconsistent, however, the correct model response should be to abstain from answering.  
Prior work has studied the abstention capabilities of LLMs~\cite{yin2023large, amayuelas2023knowledge} and proposed approaches to improve abstention when presented with insufficient context~\cite{zhou2023context,slobodkin-etal-2023-curious}, but these approaches focus on general domain question-answering (e.g.,~\squad) against a narrow range of models (e.g.,~ChatGPT).  As we show in our experiments (\S\ref{sec:rq2}), general domain settings produce highly divergent results from scientific QA settings, and different model architectures also exhibit differing abilities to abstain. Further, as new LLMs continue to be developed and released, we need an extensible way to measure 
their ability to abstain from answering questions
when provided context is irrelevant or uncertain.





To address these challenges, we introduce a framework to assess LLM abstention 
in science question-answering by removing, replacing, and augmenting provided contexts to control the answerability of questions. Using this framework, we probe abstention performance of six LLMs (\llamaShort, \vicunaShort, \flanShort, \mistral, \chatgptShort, and \gpto) on one general domain QA dataset (\squad) and three scientific QA datasets (\pubmedqa, \bioasq, \qasper). Several of these datasets include unanswerable questions by design, allowing us to analyze interactions between question answerability and noisy context.
Our work aims to answer the following: (i) How well do LLMs abstain from answering questions when the correct context is \emph{not} provided? (ii) How do context perturbations impact task performance and abstention performance? And (iii) How does question type impact task performance and abstention performance?
We summarize our contributions below:
\begin{itemize}[itemsep=0pt, topsep=3pt, leftmargin=10pt]
    \item We introduce a framework to study LLM abstention for science QA. Specifically, we probe models' ability to abstain from answering questions when the correct context is not provided, and how abstention is impacted by context perturbations and question type.

    \item Using this framework, we investigate the task and abstention performance of six LLMs on four QA datasets, ranging from general-domain factoid QA to context-sensitive, document-based science QA (\S\ref{sec:results}). Our results show that no model consistently abstains in all settings where abstention is expected (unanswerable questions and no context/random context settings), though 
    some models demonstrate a stronger ability to
    abstain for more context-dependent QA tasks, and instruction-tuned models are generally better at following abstention instructions.
    
    \item We investigate the impacts of context perturbations (\S\ref{sec:rq3}) and question type (\S\ref{sec:rq2}) on task accuracy and abstention ability. Substituting or augmenting context with random irrelevant context consistently facilitates better abstention performance across models and datasets, which can reflect as a counter-intuitive \emph{improvement} in task performance. We also find that yes-no questions tend to interfere with all models' ability to abstain relative to other question types. 
\end{itemize}

%% file: sections/related_work_new.tex
\section{Related Work}

\paragraph{Prompting LLMs for Science QA} The few-shot capability of LLMs have been applied successfully to knowledge-intensive tasks like question answering~\citep{wei2021finetuned, chowdhery2022palm,nori2023can}. Prompting strategies such as chain-of-thought~\citep{wei2022chain}, least-to-most~\citep{zhou2022least}, and others~\citep{kojima2022large,wang2022self} significantly improve LLMs' zero or few-shot abilities on diverse QA benchmarks in the general~\cite{rajpurkar-etal-2018-know, yang-etal-2015-wikiqa} and scientific~\cite{taylor2022galactica, pereira2023visconde} domains. \citet{pereira2023visconde} adopt 
retrieve-than-read on the \qasper dataset~\cite{dasigi2021dataset}, showing that current retrievers are the main bottleneck and readers (LLMs) are already performing at the human level.
Inspired by this finding, we study how irrelevant or incorrect contexts (mimicking retrieval errors) can impact LLM performance on context-intensive QA tasks. 

\paragraph{Abstention in LLMs} \citet{liao2022ptau} introduced a prompt-tuning strategy to enhance performance on unanswerable questions by mapping questions to specific templates. Beyond prompting strategies, \citet{cole-etal-2023-selectively} demonstrated that sampling methods are more effective at modulating uncertainty compared to relying on the model's likelihood. Other approaches, such as data augmentation, have been used to steer models toward abstention \cite{zhu-etal-2019-learning}. \citet{asai-choi-2021-challenges} provided a detailed analysis of language models' abstention capabilities, identifying paragraph selection and answerability prediction as key areas for improvement. Recent work has introduced new datasets to investigate whether LLMs are aware of their limitations \cite{yin2023large, amayuelas2023knowledge}, while \citet{slobodkin-etal-2023-curious} revealed that differences in LLM hidden states can delineate the boundary between known and unknown information. Abstention is also closely linked to out-of-distribution (OOD) detection and selective generation. OOD scores derived from conditional language models \cite{ren2022out} or task representations \cite{chen-etal-2023-fine} have been employed to abstain from generating low-quality outputs. In this work, we systematically characterize LLM abstention capabilities in context-dependent science question-answering tasks.


\paragraph{Context Perturbation}  Prior work studying input perturbations for NLP tasks include approaches such as model-agnostic input transformations~\cite{liang2022holistic, ravichander2022condaqa,Giorgi2022OpenDM} and adversarial example generation~\cite{jia2017adversarial, wang2021adversarial}. \citet{liang2022holistic} use semantics-preserving and semantics-altering perturbations in their robustness evaluation of LLMs. Pretrained language models can be negatively impacted by irrelevant context~\citep{chowdhery2022palm, liang2022holistic}, e.g., \citet{shi2023large} injected irrelevant numerical context for the MathQA dataset, after which ChatGPT performance dramatically decreased.
However, \citet{liang2022holistic} evaluated T5~\cite{raffel2020exploring} and PaLM~\cite{chowdhery2022palm},
demonstrating that finetuning these models with counterfactual and irrelevant contexts can improve model robustness to noisy context. Additionally, \citet{liu2023recall} created a benchmark to systematically assess LLMs' resilience against counterfactual knowledge derived from external texts, while \citet{zou2024poisonedrag} proposed an attack method that introduces malicious texts into the knowledge base of a RAG system to manipulate the LLM into generating an attacker-specified response.
In our framework, we leverage context perturbations to investigate the LLM abstention behavior for science QA, finding occasional counter-intuitive interactions between abstention and task performance.

%% file: sections/method_new.tex
\begin{table}[t!]
\centering
\small
\begin{tabular}{p{13mm}L{11mm}p{9mm}p{15mm}p{7mm}}
\toprule
 Dataset & Context length (words) & Unans. proportion & Answer types & Test set size \\ 
\midrule 
\squad  &   128  & 0.5 & Ext & 11873\\
\pubmedqa   &  204 & 0.1 & Bool &  500\\
\bioasq  & 221 & 0.0 & Bool & 140\\
\qasper  & 149 & 0.1 & Ext/Abs/Bool & 1451\\
\bottomrule
\end{tabular}
\vspace{-2mm}
  \caption{QA Dataset statistics.}
\label{tab:datasets}
\end{table}

\input{tables/unanswerable_questions}

\section{Datasets}
\label{sec:datasets}
We conduct experiments on four QA datasets: \squad~\cite{rajpurkar-etal-2018-know}, \pubmedqa~\cite{jin-etal-2019-pubmedqa}, \bioasq~\cite{nentidis2021overview}, and \qasper~\cite{dasigi2021dataset}. Dataset statistics are provided in Table~\ref{tab:datasets}. These datasets span general and science domains, and include extractive, abstractive, and boolean questions.

\begin{itemize}[itemsep=0pt, topsep=3pt, leftmargin=10pt]

\item \squad is a general-domain reading comprehension QA dataset. Answer contexts are extracted from Wikipedia. 
\item \pubmedqa is a biomedical QA dataset. Questions are automatically derived from PubMed paper titles and are answered from the conclusion sections of the corresponding abstracts. All questions can be answered Yes/No/Maybe. 
\item \bioasq includes Yes/No questions that are formulated by biomedical experts, reflecting real-life information needs encountered during their work. Answers are provided by medical experts from paper abstracts.  
\item \qasper is a full document science QA dataset. Questions are written by domain experts and answers are annotated from the full text of associated computer science papers. Questions can be boolean, extractive, or abstractive, and multiple answers may be provided for each question.

\end{itemize}

\paragraph{Unanswerable questions}  Three of these datasets contain unanswerable questions (proportions in Table~\ref{tab:datasets}, examples in Table~\ref{tab:datasets_examples}). \squad introduced unanswerable questions in machine reading comprehension;
these unanswerable questions were curated by altering questions through negation, antonym swaps, entity swaps, mutual exclusion, impossible conditions, and other ways which make it such that the context paragraph no longer implies any answer. In \squad, ``unanswerable'' questions imply \emph{irrelevant} context passages. For \pubmedqa, questions can be answered ``yes'', ``no'', or ``maybe'', and we interpret ``maybe''  as ``unanswerable'';
therefore, ``unanswerable'' in \pubmedqa can be interpreted as answers with \emph{high uncertainty} based on the given context. For \qasper, ``unanswerable'' questions are expert-labeled, and mean that \emph{no answer is available} in the given document.

\section{Framework}
We describe our framework in terms of prompting strategies, context perturbation methods, the choice of models and handling of model output, and evaluation metrics.

\subsection{Prompting strategies} 

We adopt prompting templates that achieved the best performance based on recent work~\cite{pereira2023visconde}. We refer to these templates as \textit{constrained prompts} since answer constraints (e.g., ``Answer `Yes' or `No' for boolean questions'') are added to achieve better task performance. For datasets that do not have boolean questions (e.g., \squad), we do not include boolean answer constraints. The example prompt used for the \qasper dataset is the following: \\ [-2mm]

\colorbox{blue!8}{
\begin{minipage}{0.42\textwidth}
\scriptsize
Create an Answer  to the Question using following documents. {\color{blue}{\textbf{Pay attention to only answer ``Yes'' or ``No'' for boolean questions.}}} \textbf{Answer ``Unanswerable'' when you are not sure about the answer.} \\ [-2mm]

Context: \{$c$\} \\ [-2mm]

Question: \{$q$\} \\ [-2mm]

Answer:
\end{minipage}} \\


\noindent Prompts for other datasets are in Appendix \ref{app:prompt_templates}.


We conduct zero-shot experiments. Given an input pair $(c, q)$ where $c$ is context and $q$ is question, we prepend the 
constrained prompt instructions along
with an explicit directive for handling unanswerable questions. Given complex interactions between model architecture, pretraining, instruction-tuning, dataset, question type, question answerability, context perturbations, and abstention, these settings confer significant complexity for our analysis. We therefore reserve the analysis of few-shot experiments and in-context learning to future work.  
We also conduct ablations with different prompting templates and abstention representations (results for these experiments in Appendix \ref{app:ablation}).


\subsection{Context Perturbation}
\label{sec:Context_design}

We conduct experiments to assess model sensitivity to context perturbations. We either provide the given context, or perturb the context by removing, replacing, or adding context passages (Figure~\ref{fig:framework}).

\begin{itemize}[itemsep=0pt, topsep=2pt, leftmargin=0pt]
    \item[] \textbf{Given context}: We use the original/gold context provided by the dataset. 
    For unanswerable questions in \squad, given context is unchanged but the designers manually modified the question to render the context ineffective for inferring an answer.  
    For \pubmedqa, we label a question unanswerable if the given answer is "maybe" and do not change the context. 
    For unanswerable questions in \qasper, given context is empty.
    
    \item[] \textbf{No context}: We remove the given context.
    \item[] \textbf{Random irrelevant context}: We replace the given context with the context from a random question in the train split.
    \item[] \textbf{Noisy context}: We \emph{append} context from a random question to the given context.
\end{itemize}

\subsection{Models}
\label{sec:models}

We conduct experiments using LLamaV2-13b-chat (\llamaShort), Vicuna1.5-13b-chat (\vicunaShort), Flan-T5-XL (\flanShort), Mistral-7B-Instruct-v0.1 (\mistral), \chatgpt (\chatgptShort), and gpt-4o-2024-05-13 (\gpto). We select these models to have reasonable representation across the following attributes:

\begin{itemize}[itemsep=0pt, topsep=1pt, leftmargin=10pt]
\item Closed API (\chatgptShort; \gpto) vs open weights (\llamaShort, \vicunaShort, \flanShort, and \mistral)
\item Encoder-decoder (\flanShort) vs decoder-only (\llamaShort, \vicunaShort, \mistral, \chatgptShort, and \gpto) architecture
\item Models having less (\llamaShort) or more (\vicunaShort) instruction tuning
\end{itemize}

\noindent For all models, we use the same hyperparameters at inference time. We set temperature to 0 and top-$p$ sampling to 1 to reduce the variability of model output (details in Appendix \ref{app:Hyperparameters}). 
\\ [-3mm]

\noindent\textbf{Post-processing model output} LLMs often generate lengthy responses, so we post-process to extract structured answers. For boolean questions, we take the first words of the output and map them to "yes," "no," or "unanswerable" based on their occurrence.

\subsection{Evaluation metrics}
\label{sec:eval}

We report both task performance (F1 or Accuracy) and abstention performance (rate of abstention) across different experimental settings. 

\paragraph{Task performance} For \squad and \qasper, we evaluate \emph{task performance} using $n$-gram F1 as reported in \citet{rajpurkar-etal-2018-know}. For \pubmedqa and \bioasq (boolean questions only),  we report Accuracy based on the original papers. Task performance is therefore comparable under different context settings, but is \emph{not} comparable across datasets. 

\paragraph{Abstention performance} We measure model \emph{abstention performance} by calculating the proportion of questions that is answered ``unanswerable.'' Ideally, models that are perfect at abstaining would be expected to have an abstention proportion of 1.0 when no or irrelevant context is provided for all questions. However, in reality this is not the case, since many questions may be context-independent or could be answered using model parametric knowledge. Abstention is therefore a quality dependent on (i) model ability to follow instructions (``Answer `Unanswerable' when you are not sure''), (ii) question context dependency (whether a question is answerable without context), and (iii) question difficulty (whether the question is answerable using model parametric knowledge). For clarity, we show task performance as plain numbers and abstention rates and deltas using (parentheses).

%% file: tables/unanswerable_questions.tex
\begin{table*}[th!]
\centering
\small
\begin{tabular}{lp{13cm}}
\toprule
 Datasets & Unanswerable Examples \\ 
\midrule 
\squad  &  Q: Who moved to Hollywood in 2004?~~C: ``.... Following the move to \textbf{Holyrood in 2004} this building was demolished. The former Midlothian County Buildings facing Parliament Square...'' \\
\pubmedqa  &  Q: Does rugby headgear prevent concussion?~~C: ``...In addition, coaches from all four levels were \textbf{questioned} about team policies and their personal opinions about the use of headgear to prevent concussion. Although the players tended to believe that the headgear could prevent concussion (62\%), the coaches were \textbf{less convinced (33\%)}...''  \\
\qasper  & Q: How many Universal Dependency features are considered?~~C: Empty. \\ 
\bottomrule
\end{tabular}
\vspace{-2mm}
  \caption{Example unanswerable questions from datasets (Q: question, C: context).}
\label{tab:datasets_examples}
\end{table*}

%% file: sections/exp_new.tex



\input{tables/task_performance}

\input{tables/abstention_combined}

\section{Results}
\label{sec:results}

\begin{figure*}[t!]
\centering
\minipage{\linewidth}
\includegraphics[width=\linewidth]{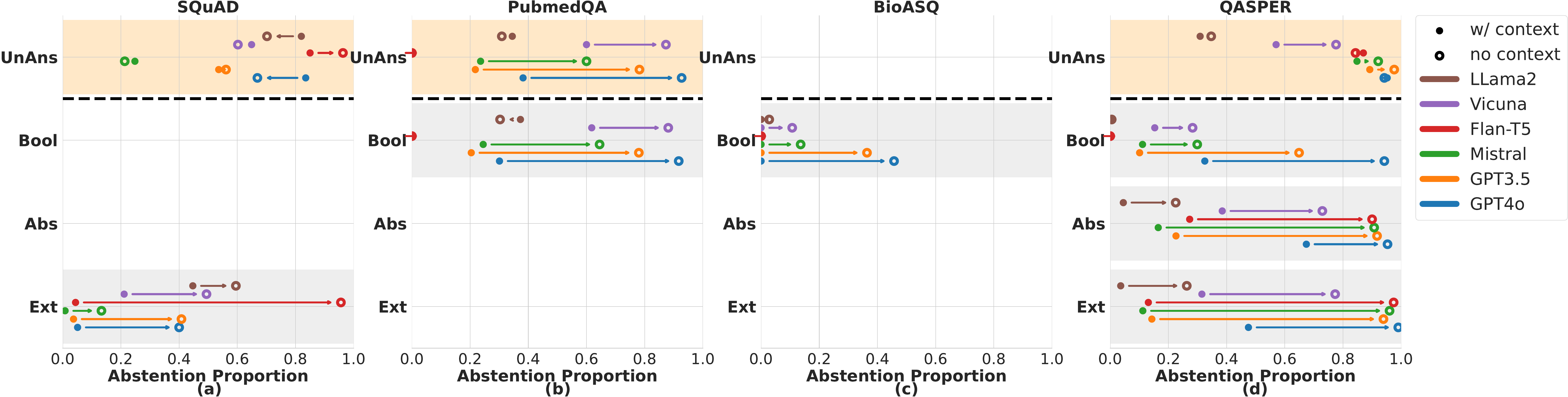}
\vspace{-6mm}
\caption{Abstention performance changes from ``with context'' to ``no context'' settings across different question types. Each row represents one question type, from top to bottom: ``Unanswerable'', ``Boolean'', ``Abstractive'' and ``Extractive''. White background indicates a dataset does not have questions of that type.}\label{fig:all_qtype}
\endminipage
\end{figure*}

We report baseline task performance (Table~\ref{tab:baseline_constrain}) and abstention performance (Table~\ref{tab:abstention_all}) for all models on all datasets. Results for context perturbations are shown as changes to task performance and abstention performance from baseline. Additional analysis by question type is shown in Figure~\ref{fig:all_qtype}. 

Baseline task performance in the zero-shot prompting setting is presented in Table~\ref{tab:baseline_constrain}(a), alongside previously reported SOTA zero-shot performance of LLMs on each dataset. Surprisingly, \flanShort achieves the best performance on \squad, \pubmedqa, and \qasper, while \llamaShort performs best on \bioasq. \chatgptShort achieves the second best performance on  \pubmedqa, \bioasq, and \qasper; and \gpto achieves second best performance on \squad, \pubmedqa, and \bioasq. Task performance of tested models are close to and sometimes exceed (\bioasq) the reported SOTA on each dataset. Though in this work, we focus on assessing change to task performance under different context settings rather than maximizing model performance.

\subsection{Impact of context perturbations}
\label{sec:rq3}

\paragraph{Random context facilitates abstention} In Table~\ref{tab:abstention_all}(c), we observe that all models are much more likely to abstain from answering when given random context except on the \bioasq dataset. For \squad and \pubmedqa, random context improves most models' abstention performance to close to 1. Specifically, random context improves \flanShort's ability to abstain from answering on \pubmedqa while the no context setting completely fails to do so. For \qasper, the abstention performance on answerable questions in the random context setting is much higher compared to no context, but is still far from 1. Oddly, for boolean questions in BioASQ, random context biases the model towards answering “no” rather than “unanswerable”, and we observe no changes in abstention rate (further analysis in Section \ref{sec:error_analysis}).

\paragraph{Adding noisy context can counter-intuitively improve task performance on some datasets} While noisy context has mixed effects on abstention (Table~\ref{tab:abstention_all}(d)), this perturbation does not always translate to negative impacts on task performance. In Table~\ref{tab:baseline_constrain}(d), task performance increases on \qasper for \llamaShort and \vicunaShort and \pubmedqa for \vicunaShort and \gpto.
For \qasper, unanswerable questions have empty context, so the baseline task performance for unanswerable questions is actually computed with no context; perturbing no context by adding noisy context therefore leads to a trade-off in task performance and abstention performance (some models abstain much more for unanswerable questions and task performance consequently improves). 

\subsection{Impact of question type}
\label{sec:rq2}

\paragraph{Models' abstention capabilities vary by question type} For extractive questions, Figure~\ref{fig:all_qtype}(a) and (d) show that model abstention varies significantly between given context and no context settings; under no context settings, models can achieve abstention performance close to 1 for a highly context-dependent dataset like \qasper. Abstractive answers show similar patterns. 
Notably, we find \emph{all} models are reluctant to abstain from boolean questions for all datasets; this overconfidence is apparent even for QASPER, which is a highly document-specific QA dataset. Surprisingly, \flanShort and \llamaShort demonstrate near zero abstention performance on boolean questions on any dataset, regardless of whether or not context is provided.\footnote{We conduct prompt ablations in Appendix~\ref{app:ablation}, showing the lack of abstention on boolean questions is highly sensitive to prompt wording. When a boolean instruction is removed from the prompt, both models are able to abstain much better for boolean questions.} \vicunaShort and \mistral perform slightly better, with some variability across datasets. \chatgptShort and \gpto demonstrate much higher abstention performance compared to other models, but are still far from 1.


\paragraph{Abstention performance on unanswerable questions varies by dataset}
In Figure~\ref{fig:all_qtype}, we examine the abstention performance on unanswerable questions (yellow background) and find that performance varies significantly across datasets and models. For \squad: \llamaShort, \vicunaShort, \mistral, and \gpto abstain less when context is removed, which contradicts intuition. In \squad, unanswerable questions are still related to the provided context; which may explain why when the confusing context is removed, these two models become more likely to answer. \chatgptShort performs consistently regardless of whether context is provided. This consistency suggests that \chatgptShort may rely on internal mechanisms for abstention that are less sensitive to the presence or absence of contextual information.
For \pubmedqa, unanswerable questions are associated with context that is highly uncertain; \flanShort and \llamaShort consistently refuse to abstain from answering these questions, which are boolean, while \vicunaShort follows the ``unanswerable'' instruction quite well.
For \qasper, abstention performance is  generally higher across all models compared to \squad and \pubmedqa since the questions are more document-grounded.


\begin{table}[t]
\centering
\small
\begin{tabular}{lcccc}
\toprule
 Model &  \scriptsize \squad & \scriptsize \pubmedqa & \scriptsize \bioasq & \scriptsize \qasper \\ 
\midrule
 \llamaShort & 29.8 & 16.6 &  2.8 & 20.6   \\
 \vicunaShort & 33.6& 29.0 & 10.7 & 44.9 \\
\flanShort & \textbf{91.2}  & 0.0 & 0.0 & \underline{66.0} \\
 \mistral & 12.6  & 43.2 & 13.6 & 64.5 \\
 \chatgptShort & \underline{38.8} & \underline{59.3} & \underline{36.4} & \textbf{75.7}\\
 \gpto & 22.5 & \textbf{62.0} & \textbf{45.7} & 38.2\\
\bottomrule
\end{tabular}
  \caption{Percentage of answerable questions for which the model changes from answering to abstaining when context is removed. Best performances \textbf{bolded}, second best underlined.
  }
\label{tab:hasans_to_unans}
\end{table}

\subsection{Abstention performance by model}
\label{sec:rq1}




\paragraph{Instruction-tuned models more readily abstain} Table~\ref{tab:hasans_to_unans} shows the percentage of answers for which the model changed from answering to abstaining when context is removed. The expectation for context-dependent questions would be that a model would abstain from answering if no context is provided. \gpto achieves the largest abstention changes on \pubmedqa and \bioasq, while \chatgptShort also abstain well across all four datasets, indicating that they are responsive to context loss. \vicunaShort, with more instruction-tuning than \llamaShort, demonstrates relatively strong abstention behavior in these cases compared to \llamaShort.


\paragraph{Flan-T5 readily abstains for non-boolean questions} \flanShort abstains well when context is removed (Table~\ref{tab:hasans_to_unans}). 
We observe large changes for \squad and \qasper (comparable or better than \gpto and \chatgptShort), although this behavior does not generalize \emph{at all} to boolean questions---the change is 0.0 for \pubmedqa and \bioasq, which consist only of boolean questions. Different prompting strategies may be necessary to enjoin certain models to abstain when answering boolean questions.


\input{tables/cases_abstention}

\section{Error Analysis} \label{sec:error_analysis}
We perform error analysis on a consistently high performing model in our analysis, \chatgptShort, and summarize main reasons why the model may fail to abstain. We sample 20 failure cases each from among answerable questions and unanswerable questions, for both gold context and no context settings. We present our findings by dataset.

\begin{itemize}[itemsep=0pt, topsep=2pt, leftmargin=0pt]
    \item[] \textbf{\squad}~~ 
    Answerable questions are either general (e.g., ``What German poet was descended from Huguenots?'') or could be answered without context (e.g., ``What types of pumps are typically used in industrial boilers?''); the model tends to answer these regardless of whether context is provided.
    For unanswerable questions, half of failure cases are very open-ended like ``What effect do technologies and resources generate?''~and removing context will cause \chatgptShort to answer rather than abstain. Another 20\% of questions contain popular entities but have no correct answer, e.g., ``What lake contains part of the Rhine Falls?'' misleads the model to generate ``Lake Constance'' as an answer since ``The Rhine emerges from Lake Constance.'' 

    
    \item[] \textbf{\pubmedqa}~~\chatgptShort is strongly inclined to answer ``yes'' for all failure cases among both answerable and unanswerable questions under the no context setting, perhaps due to the phrasing of automatically constructed questions.
   
    \item[] \textbf{\bioasq}~~ Around 90\% of failure cases we sampled were answered correctly by \chatgptShort without context, including both ``yes'' and ``no'' answers. 
    Another 10\% of cases are affected by the model's tendency to hallucinate ``yes'', which is consistent with cases in \pubmedqa. We manually substitute antonyms for the word in brackets on failure cases such as ``Is Apelin usually [decreased] in diabetes?'', ``Does vesatolimod [inhibit] TLR7?'' etc., and \chatgptShort still always answers ``yes''. Interestingly, for these factoid questions, substituting gold with random context skews the model strongly towards answering ``no''. These two types of hallucination behavior require further investigation. Our observations also raise concerns for assessing LLM performance using QA benchmarks containing boolean questions. 
    
    \item[] \textbf{\qasper}~~ For answerable questions, 50\% of failures seem to be caused by \chatgptShort's ignorance of the ambiguity resulting from anaphora; for instance, \chatgptShort should not be able to resolve terms in questions such as ``they'',``this study'', and ``the models'' without context, yet the model answers these questions anyway under the no context setting. Around 30\% of questions are very general, such as 
    ``What is a soft label?'' and ``Why is NER for tweets more challenging as the number of entities increases?''---these questions lead the model to answer rather than abstain. For unanswerable questions, the distribution of failure reasons is similar.
\end{itemize}

\noindent Table~\ref{tab:case_study} shows two context perturbation examples with answers generated by \chatgptShort. More examples are given in Appendix \ref{app:error_analysis}.

%% file: tables/task_performance.tex
\begin{table}[t!]
\centering
\small
\setlength{\tabcolsep}{5pt}
\begin{tabular}{lcccc}
\toprule
\scriptsize &  \scriptsize \squad & \scriptsize \pubmedqa & \scriptsize \bioasq & \scriptsize \qasper \\
\midrule
\textsc{SOTA}  & 90.5* & 77.6** &  94.3** & 61.4$\dagger$  \\
\midrule
\multicolumn{5}{c}{(a) Given context} \\
\midrule
\llamaShort & 51.7 & 52.6 &  \textbf{98.6} & 16.8   \\
 \vicunaShort & 61.0 & 36.4 & 93.6 & 30.5\\
\flanShort & \textbf{87.4} & \textbf{73.2}  & 97.8  & \textbf{60.1} \\
 \mistral & 38.5 & 58.4  & 94.2  & 50.9 \\
 \chatgptShort & 60.4 & \underline{61.2} & \underline{97.9} & \underline{57.8} \\
 \gpto & \underline{82.6} & \underline{61.2} & \underline{97.9} & 54.1 \\
\midrule
\multicolumn{5}{c}{(b) No context} \\
\midrule
 \llamaShort &  \midpointgradientcell{-11.7}{-46.4}{20}{0}{neg}{pos}{\opacity}{0} & \midpointgradientcell{-17.8}{-46.4}{20}{0}{neg}{pos}{\opacity}{0} & \midpointgradientcell{-43.6}{-46.4}{20}{0}{neg}{pos}{\opacity}{0} & \midpointgradientcell{-2.7 }{-46.4}{20}{0}{neg}{pos}{\opacity}{0}   \\
\vicunaShort & \midpointgradientcell{-24.0}{-46.4}{20}{0}{neg}{pos}{\opacity}{0} & \midpointgradientcell{-19.4}{-46.4}{20}{0}{neg}{pos}{\opacity}{0} & \midpointgradientcell{-32.9}{-46.4}{20}{0}{neg}{pos}{\opacity}{0} & \midpointgradientcell{-10.4}{-46.4}{20}{0}{neg}{pos}{\opacity}{0}\\
 \flanShort & \midpointgradientcell{-38.2}{-46.4}{20}{0}{neg}{pos}{\opacity}{0} & \midpointgradientcell{-16.4}{-46.4}{20}{0}{neg}{pos}{\opacity}{0}   & \midpointgradientcell{-30.8}{-46.4}{20}{0}{neg}{pos}{\opacity}{0}  & \midpointgradientcell{-38.0}{-46.4}{20}{0}{neg}{pos}{\opacity}{0}  \\
  \mistral & \midpointgradientcell{-20.9}{-46.4}{20}{0}{neg}{pos}{\opacity}{0} & \midpointgradientcell{-31.0}{-46.4}{20}{0}{neg}{pos}{\opacity}{0}   & \midpointgradientcell{-32.1}{-46.4}{20}{0}{neg}{pos}{\opacity}{0}  & \midpointgradientcell{-29.4}{-46.4}{20}{0}{neg}{pos}{\opacity}{0}  \\
  
\chatgptShort & \midpointgradientcell{-22.9}{-46.4}{20}{0}{neg}{pos}{\opacity}{0}  & \midpointgradientcell{-39.0}{-46.4}{20}{0}{neg}{pos}{\opacity}{0}  & \midpointgradientcell{-46.4}{-46.4}{20}{0}{neg}{pos}{\opacity}{0} & \midpointgradientcell{-37.7}{-46.4}{20}{0}{neg}{pos}{\opacity}{0}\\

\gpto & \midpointgradientcell{-36.0}{-46.4}{20}{0}{neg}{pos}{\opacity}{0}  & \midpointgradientcell{-45.6}{-46.4}{20}{0}{neg}{pos}{\opacity}{0}  & \midpointgradientcell{-49.3}{-46.4}{20}{0}{neg}{pos}{\opacity}{0} & \midpointgradientcell{-35.6}{-46.4}{20}{0}{neg}{pos}{\opacity}{0}\\
\midrule
\multicolumn{5}{c}{(c) Random context} \\
\midrule
 \llamaShort &  \midpointgradientcell{-1.9}{-64.9}{1.1}{0}{neg}{pos}{\opacity}{0} & \midpointgradientcell{-40.8}{-64.9}{20}{0}{neg}{pos}{\opacity}{0} & \midpointgradientcell{-63.6}{-64.9}{20}{0}{neg}{pos}{\opacity}{0} & \midpointgradientcell{-0.8 }{-64.9}{1.1}{0}{neg}{pos}{\opacity}{0}   \\
\vicunaShort & \midpointgradientcell{-11.8}{-64.9}{20}{0}{neg}{pos}{\opacity}{0} & \midpointgradientcell{-25.2}{-64.9}{20}{0}{neg}{pos}{\opacity}{0} & \midpointgradientcell{-41.5}{-64.9}{20}{0}{neg}{pos}{\opacity}{0} & \midpointgradientcell{-11.5}{-64.9}{20}{0}{neg}{pos}{\opacity}{0}\\
 \flanShort & \midpointgradientcell{-37.3}{-64.9}{20}{0}{neg}{pos}{\opacity}{0} & \midpointgradientcell{-59.4}{-64.9}{20}{0}{neg}{pos}{\opacity}{0}   & \midpointgradientcell{-63.5}{-64.9}{20}{0}{neg}{pos}{\opacity}{0}  & \midpointgradientcell{-37.2}{-64.9}{20}{0}{neg}{pos}{\opacity}{0}  \\
  \mistral & \midpointgradientcell{4.6}{-46.4}{20}{0}{neg}{pos}{\opacity}{0} & \midpointgradientcell{-47.4}{-46.4}{20}{0}{neg}{pos}{\opacity}{0}   & \midpointgradientcell{-59.2}{-46.4}{20}{0}{neg}{pos}{\opacity}{0}  & \midpointgradientcell{-30.5}{-46.4}{20}{0}{neg}{pos}{\opacity}{0}  \\
  
\chatgptShort & \midpointgradientcell{-10.4}{-64.9}{20}{0}{neg}{pos}{\opacity}{0}  & \midpointgradientcell{-50.0}{-64.9}{20}{0}{neg}{pos}{\opacity}{0}  & \midpointgradientcell{-64.9}{-64.9}{20}{0}{neg}{pos}{\opacity}{0} & \midpointgradientcell{-37.6}{-64.9}{20}{0}{neg}{pos}{\opacity}{0}\\

\gpto & \midpointgradientcell{-32.6}{-46.4}{20}{0}{neg}{pos}{\opacity}{0}  & \midpointgradientcell{-51.2}{-46.4}{20}{0}{neg}{pos}{\opacity}{0}  & \midpointgradientcell{-64.3}{-46.4}{20}{0}{neg}{pos}{\opacity}{0} & \midpointgradientcell{-35.0}{-46.4}{20}{0}{neg}{pos}{\opacity}{0}\\

\midrule
\multicolumn{5}{c}{(d) Noisy context} \\
\midrule
\llamaShort &\midpointgradientcell{-4.3}{-13.2}{4.8}{0}{neg}{pos}{\opacity}{0}& \midpointgradientcell{-13.2}{-13.2}{4.8}{0}{neg}{pos}{\opacity}{0} & \midpointgradientcell{-10.7}{-13.2}{4.8}{0}{neg}{pos}{\opacity}{0} & \midpointgradientcell{1.9}{-13.2}{4.8}{0}{neg}{pos}{\opacity}{0}  \\
\vicunaShort & \midpointgradientcell{-1.4}{-13.2}{4.8}{0}{neg}{pos}{\opacity}{0}& \midpointgradientcell{4.8}{-13.2}{4.8}{0}{neg}{pos}{\opacity}{0} & \midpointgradientcell{-0.7}{-13.2}{4.8}{0}{neg}{pos}{\opacity}{0} & \midpointgradientcell{4.7}{-13.2}{4.8}{0}{neg}{pos}{\opacity}{0}  \\
\flanShort & \midpointgradientcell{0.0}{-13.2}{4.8}{0}{neg}{pos}{\opacity}{0}  & \midpointgradientcell{-0.2}{-13.2}{4.8}{0}{neg}{pos}{\opacity}{0} & \midpointgradientcell{-0.7}{-13.2}{4.8}{0}{neg}{pos}{\opacity}{0} & \midpointgradientcell{0.0}{-13.2}{4.8}{0}{neg}{pos}{\opacity}{0}\\

 \mistral & \midpointgradientcell{-2.6}{-46.4}{20}{0}{neg}{pos}{\opacity}{0} & \midpointgradientcell{-8.2}{-46.4}{20}{0}{neg}{pos}{\opacity}{0}   & \midpointgradientcell{-2.8}{-46.4}{20}{0}{neg}{pos}{\opacity}{0}  & \midpointgradientcell{0.0}{-46.4}{20}{0}{neg}{pos}{\opacity}{0}  \\
 
\chatgptShort & \midpointgradientcell{-2.4}{-13.2}{4.8}{0}{neg}{pos}{\opacity}{0} & \midpointgradientcell{-2.8}{-13.2}{4.8}{0}{neg}{pos}{\opacity}{0} & \midpointgradientcell{-2.2}{-13.2}{4.8}{0}{neg}{pos}{\opacity}{0}& \midpointgradientcell{-2.4}{-13.2}{4.8}{0}{neg}{pos}{\opacity}{0}\\

\gpto & \midpointgradientcell{-0.6}{-46.4}{20}{0}{neg}{pos}{\opacity}{0}  & \midpointgradientcell{1.4}{-46.4}{20}{0}{neg}{pos}{\opacity}{0}  & \midpointgradientcell{-0.8}{-46.4}{20}{0}{neg}{pos}{\opacity}{0} & \midpointgradientcell{-5.2}{-46.4}{20}{0}{neg}{pos}{\opacity}{0}\\

\bottomrule
\end{tabular}
  \caption{Model zero-shot task performance using constrained prompts. SOTA indicates the best zero-shot performance of LLMs reported in previous papers (*=Flan-UL2~\cite{slobodkin-etal-2023-curious}, **=Galactica~\cite{taylor2022galactica}), or best performance from a pretrained LM ($\dagger$=UnifiedQA-large~\cite{dasigi2021dataset}). (a) Among baseline model performance with the given context, the best performances are \textbf{bolded}, second best \underline{underlined}. Colors indicate \hlpos{positive} or \hlneg{negative} delta from baseline task performance with different context perturbations: (b) no context, (c) random context, and (d) noisy context. While task performance generally degrades with context perturbations, this is not consistently the case for \qasper due to interactions between abstention and task performance (see Section~\ref{sec:rq3}).}

\label{tab:baseline_constrain}
\end{table}

\vspace{-4mm}

%% file: tables/abstention_combined.tex
\begin{table*}[t!]
\centering
\small
\begin{tabular}{c|cc|cc|c|cc}
\toprule
Baseline & \multicolumn{2}{c}{SQUAD2} & \multicolumn{2}{c}{Pubmed} & \multicolumn{1}{c}{BioASQ} & \multicolumn{2}{c}{Qasper} \\
\midrule
Model & Ans. & Unans. & Ans. & Unans. & Ans. & Ans. & Unans. \\
\midrule
\llamaShort & (\baseline{44.7}) & (\baseline{82.0}) & (\baseline{37.3}) & (\baseline{34.5}) & (\baseline{0.0}) & (\baseline{3.1}) & (\baseline{30.9}) \\
\vicunaShort & (\baseline{21.1}) & (\baseline{64.9}) & (\baseline{61.8}) & (\baseline{60.0}) & (\baseline{0.0}) & (\baseline{30.0}) & (\baseline{57.0}) \\
\flanShort & (\baseline{4.4}) & (\baseline{85.0}) & (\baseline{0.0}) & (\baseline{0.0}) & (\baseline{0.0}) & (\baseline{13.0}) & (\baseline{87.0}) \\

\mistral & (\baseline{	0.8}) & (\baseline{24.9}) & (\baseline{24.5}) &  (\baseline{23.6})& (\baseline{0.0}) & (\baseline{12.1}) & (\baseline{84.8})   \\
\chatgptShort & (\baseline{3.7}) & (\baseline{53.3}) & (\baseline{20.4}) & (\baseline{21.8}) & (\baseline{0.0}) & (\baseline{15.0}) & (\baseline{89.2}) \\

\gpto & (\baseline{5.1}) & (\baseline{83.5}) & (\baseline{30.1}) & (\baseline{38.2}) & (\baseline{0.0}) & (\baseline{41.1}) & (\baseline{98.6}) \\

\midrule
\multicolumn{8}{c}{(b) No context} \\
\midrule
\llamaShort & (\midpointgradientcell{14.8}{-13}{80}{2.9}{neg}{pos}{\opacity}{0}) & (\midpointgradientcell{-11.7}{-13}{80}{0}{neg}{pos}{\opacity}{0}) & (\midpointgradientcell{-7.0}{-13}{80}{0}{neg}{pos}{\opacity}{0}) & (\midpointgradientcell{-3.6}{-13}{80}{0}{neg}{pos}{\opacity}{0}) & (\midpointgradientcell{2.8}{-13}{80}{0}{neg}{pos}{\opacity}{0}) & (\midpointgradientcell{17.7}{-13}{80}{0}{neg}{pos}{\opacity}{0}) & (\midpointgradientcell{3.7}{-13}{80}{0}{neg}{pos}{\opacity}{0}) \\
\vicunaShort & (\midpointgradientcell{28.3}{-13}{80}{0}{neg}{pos}{\opacity}{0}) & (\midpointgradientcell{-4.7}{-13}{80}{0}{neg}{pos}{\opacity}{0}) & (\midpointgradientcell{26.7}{-13}{80}{0}{neg}{pos}{\opacity}{0}) & (\midpointgradientcell{27.3}{-13}{80}{0}{neg}{pos}{\opacity}{0}) & (\midpointgradientcell{10.7}{-13}{80}{0}{neg}{pos}{\opacity}{0}) & (\midpointgradientcell{38.2}{-13}{80}{0}{neg}{pos}{\opacity}{0}) & (\midpointgradientcell{20.7}{-13}{80}{0}{neg}{pos}{\opacity}{0}) \\
\flanShort & (\midpointgradientcell{91.2}{-13}{80}{0}{neg}{pos}{\opacity}{0}) & (\midpointgradientcell{11.3}{-13}{80}{0}{neg}{pos}{\opacity}{0}) & (\midpointgradientcell{0.0}{-13}{80}{0}{neg}{pos}{\opacity}{0}) & (\midpointgradientcell{0.0}{-13}{80}{0}{neg}{pos}{\opacity}{0}) & (\midpointgradientcell{0.0}{-13}{80}{0}{neg}{pos}{\opacity}{0}) & (\midpointgradientcell{65.0}{-13}{80}{0}{neg}{pos}{\opacity}{0}) & (\midpointgradientcell{-2.8}{-13}{80}{0}{neg}{pos}{\opacity}{0}) \\

\mistral & (\midpointgradientcell{12.5}{-13}{80}{0}{neg}{pos}{\opacity}{0}) & (\midpointgradientcell{-3.6}{-13}{80}{0}{neg}{pos}{\opacity}{0}) & (\midpointgradientcell{40.0}{-13}{80}{0}{neg}{pos}{\opacity}{0}) & (\midpointgradientcell{36.4}{-13}{80}{0}{neg}{pos}{\opacity}{0}) & (\midpointgradientcell{13.5	}{-13}{80}{0}{neg}{pos}{\opacity}{0}) & (\midpointgradientcell{71.5}{-13}{80}{0}{neg}{pos}{\opacity}{0}) & (\midpointgradientcell{7.3}{-13}{80}{0}{neg}{pos}{\opacity}{0}) \\

\chatgptShort & (\midpointgradientcell{37.1}{-13}{80}{0}{neg}{pos}{\opacity}{0}) & (\midpointgradientcell{2.5}{-13}{80}{0}{neg}{pos}{\opacity}{0}) & (\midpointgradientcell{57.6}{-13}{80}{0}{neg}{pos}{\opacity}{0}) & (\midpointgradientcell{56.4}{-13}{80}{0}{neg}{pos}{\opacity}{0}) & (\midpointgradientcell{36.4}{-13}{80}{0}{neg}{pos}{\opacity}{0}) & (\midpointgradientcell{73.8}{-13}{80}{0}{neg}{pos}{\opacity}{0}) & (\midpointgradientcell{8.4}{-13}{80}{0}{neg}{pos}{\opacity}{0}) \\

\gpto & (\midpointgradientcell{34.9}{-13}{80}{0}{neg}{pos}{\opacity}{0}) & (\midpointgradientcell{-16.6}{-13}{80}{0}{neg}{pos}{\opacity}{0}) & (\midpointgradientcell{61.6}{-13}{80}{0}{neg}{pos}{\opacity}{0}) & (\midpointgradientcell{53.5}{-13}{80}{0}{neg}{pos}{\opacity}{0}) & (\midpointgradientcell{45.7}{-13}{80}{0}{neg}{pos}{\opacity}{0}) & (\midpointgradientcell{56.4}{-13}{80}{0}{neg}{pos}{\opacity}{0}) & (\midpointgradientcell{0.5}{-13}{80}{0}{neg}{pos}{\opacity}{0}) \\

\midrule
\multicolumn{8}{c}{(c) Random context} \\
\midrule
\llamaShort & (\midpointgradientcell{54.8}{-13}{80}{0}{neg}{pos}{\opacity}{0}) & (\midpointgradientcell{17.2}{-13}{80}{0}{neg}{pos}{\opacity}{0}) & (\midpointgradientcell{58.7}{-13}{80}{0}{neg}{pos}{\opacity}{0}) & (\midpointgradientcell{60.0}{-13}{80}{0}{neg}{pos}{\opacity}{0}) & (\midpointgradientcell{0.0}{-13}{80}{0}{neg}{pos}{\opacity}{0}) & (\midpointgradientcell{16.9}{-13}{80}{0}{neg}{pos}{\opacity}{0}) & (\midpointgradientcell{17.8}{-13}{80}{0}{neg}{pos}{\opacity}{0}) \\
\vicunaShort & (\midpointgradientcell{73.8}{-13}{80}{0}{neg}{pos}{\opacity}{0}) & (\midpointgradientcell{31.7}{-13}{80}{0}{neg}{pos}{\opacity}{0}) & (\midpointgradientcell{37.5}{-13}{80}{0}{neg}{pos}{\opacity}{0}) & (\midpointgradientcell{40.0}{-13}{80}{0}{neg}{pos}{\opacity}{0}) & (\midpointgradientcell{0.0}{-13}{80}{0}{neg}{pos}{\opacity}{0}) & (\midpointgradientcell{44.0}{-13}{80}{0}{neg}{pos}{\opacity}{0}) & (\midpointgradientcell{34.5}{-13}{80}{0}{neg}{pos}{\opacity}{0}) \\
\flanShort & (\midpointgradientcell{95.4}{-13}{80}{0}{neg}{pos}{\opacity}{0}) & (\midpointgradientcell{15.0}{-13}{80}{0}{neg}{pos}{\opacity}{0}) & (\midpointgradientcell{91.5}{-13}{80}{0}{neg}{pos}{\opacity}{0}) & (\midpointgradientcell{92.7}{-13}{80}{0}{neg}{pos}{\opacity}{0}) & (\midpointgradientcell{0.0}{-13}{80}{0}{neg}{pos}{\opacity}{0}) & (\midpointgradientcell{54.6}{-13}{80}{0}{neg}{pos}{\opacity}{0}) & (\midpointgradientcell{-0.9}{-13}{80}{0}{neg}{pos}{\opacity}{0}) \\

\mistral & (\midpointgradientcell{83.3}{-13}{80}{0}{neg}{pos}{\opacity}{0}) & (\midpointgradientcell{59.3}{-13}{80}{0}{neg}{pos}{\opacity}{0}) & (\midpointgradientcell{75.5}{-13}{80}{0}{neg}{pos}{\opacity}{0}) & (\midpointgradientcell{76.4}{-13}{80}{0}{neg}{pos}{\opacity}{0}) & (\midpointgradientcell{0.0}{-13}{80}{0}{neg}{pos}{\opacity}{0}) & (\midpointgradientcell{58.1}{-13}{80}{0}{neg}{pos}{\opacity}{0}) & (\midpointgradientcell{6.8}{-13}{80}{0}{neg}{pos}{\opacity}{0}) \\

\chatgptShort & (\midpointgradientcell{95.5}{-13}{80}{0}{neg}{pos}{\opacity}{0}) & (\midpointgradientcell{45.9}{-13}{80}{0}{neg}{pos}{\opacity}{0}) & (\midpointgradientcell{79.6}{-13}{80}{0}{neg}{pos}{\opacity}{0}) & (\midpointgradientcell{78.2}{-13}{80}{0}{neg}{pos}{\opacity}{0}) & (\midpointgradientcell{0.0}{-13}{80}{0}{neg}{pos}{\opacity}{0}) & (\midpointgradientcell{66.9}{-13}{80}{0}{neg}{pos}{\opacity}{0}) & (\midpointgradientcell{5.5}{-13}{80}{0}{neg}{pos}{\opacity}{0}) \\

\gpto & (\midpointgradientcell{94.9}{-13}{80}{0}{neg}{pos}{\opacity}{0}) & (\midpointgradientcell{16.3}{-13}{80}{0}{neg}{pos}{\opacity}{0}) & (\midpointgradientcell{69.9}{-13}{80}{0}{neg}{pos}{\opacity}{0}) & (\midpointgradientcell{61.8}{-13}{80}{0}{neg}{pos}{\opacity}{0}) & (\midpointgradientcell{0.0}{-13}{80}{0}{neg}{pos}{\opacity}{0}) & (\midpointgradientcell{53.2}{-13}{80}{0}{neg}{pos}{\opacity}{0}) & (\midpointgradientcell{0.9}{-13}{80}{0}{neg}{pos}{\opacity}{0}) \\
\midrule
\multicolumn{8}{c}{(d) Noisy context} \\
\midrule
\llamaShort & (\midpointgradientcell{22.8}{-13}{80}{0}{neg}{pos}{\opacity}{0}) & (\midpointgradientcell{3.2}{-13}{80}{0}{neg}{pos}{\opacity}{0}) & (\midpointgradientcell{18.7}{-13}{80}{0}{neg}{pos}{\opacity}{0}) & (\midpointgradientcell{18.2}{-13}{80}{0}{neg}{pos}{\opacity}{0}) & (\midpointgradientcell{0.0}{-13}{80}{0}{neg}{pos}{\opacity}{0}) & (\midpointgradientcell{2.2}{-13}{80}{0}{neg}{pos}{\opacity}{0}) & (\midpointgradientcell{17.4}{-13}{80}{0}{neg}{pos}{\opacity}{0}) \\
\vicunaShort & (\midpointgradientcell{8.2}{-13}{80}{0}{neg}{pos}{\opacity}{0}) & (\midpointgradientcell{2.5}{-13}{80}{0}{neg}{pos}{\opacity}{0}) & (\midpointgradientcell{-11.0}{-13}{80}{0}{neg}{pos}{\opacity}{0}) & (\midpointgradientcell{-12.7}{-13}{80}{0}{neg}{pos}{\opacity}{0}) & (\midpointgradientcell{0.0}{-13}{80}{0}{neg}{pos}{\opacity}{0}) & (\midpointgradientcell{13.2}{-13}{80}{0}{neg}{pos}{\opacity}{0}) & (\midpointgradientcell{34.7}{-13}{80}{0}{neg}{pos}{\opacity}{0}) \\
\flanShort & (\midpointgradientcell{-0.9}{-13}{80}{0}{neg}{pos}{\opacity}{0}) & (\midpointgradientcell{-1.3}{-13}{80}{0}{neg}{pos}{\opacity}{0}) & (\midpointgradientcell{0.0}{-13}{80}{0}{neg}{pos}{\opacity}{0}) & (\midpointgradientcell{0.0}{-13}{80}{0}{neg}{pos}{\opacity}{0}) & (\midpointgradientcell{0.0}{-13}{80}{0}{neg}{pos}{\opacity}{0}) & (\midpointgradientcell{-1.4}{-13}{80}{0}{neg}{pos}{\opacity}{0}) & (\midpointgradientcell{-2.7}{-13}{80}{0}{neg}{pos}{\opacity}{0}) \\

\mistral & (\midpointgradientcell{0.2}{-13}{80}{0}{neg}{pos}{\opacity}{0}) & (\midpointgradientcell{-2.4}{-13}{80}{0}{neg}{pos}{\opacity}{0}) & (\midpointgradientcell{15.7}{-13}{80}{0}{neg}{pos}{\opacity}{0}) & (\midpointgradientcell{20}{-13}{80}{0}{neg}{pos}{\opacity}{0}) & (\midpointgradientcell{0.0}{-13}{80}{0}{neg}{pos}{\opacity}{0}) & (\midpointgradientcell{2.2}{-13}{80}{0}{neg}{pos}{\opacity}{0}) & (\midpointgradientcell{6.4}{-13}{80}{0}{neg}{pos}{\opacity}{0}) \\

\chatgptShort & (\midpointgradientcell{1.5}{-13}{80}{0}{neg}{pos}{\opacity}{0}) & (\midpointgradientcell{-2.8}{-13}{80}{0}{neg}{pos}{\opacity}{0}) & (\midpointgradientcell{3.2}{-13}{80}{0}{neg}{pos}{\opacity}{0}) & (\midpointgradientcell{0.0}{-13}{80}{0}{neg}{pos}{\opacity}{0}) & (\midpointgradientcell{0.0}{-13}{80}{0}{neg}{pos}{\opacity}{0}) & (\midpointgradientcell{4.5}{-13}{80}{0}{neg}{pos}{\opacity}{0}) & (\midpointgradientcell{4.6}{-13}{80}{0}{neg}{pos}{\opacity}{0}) \\

\gpto & (\midpointgradientcell{0.0}{-13}{80}{0}{neg}{pos}{\opacity}{0}) & (\midpointgradientcell{-2.4}{-13}{80}{0}{neg}{pos}{\opacity}{0}) & (\midpointgradientcell{-4.7}{-13}{80}{0}{neg}{pos}{\opacity}{0}) & (\midpointgradientcell{-3.7}{-13}{80}{0}{neg}{pos}{\opacity}{0}) & (\midpointgradientcell{0.0}{-13}{80}{0}{neg}{pos}{\opacity}{0}) & (\midpointgradientcell{6.6}{-13}{80}{0}{neg}{pos}{\opacity}{0}) & (\midpointgradientcell{0.9}{-13}{80}{0}{neg}{pos}{\opacity}{0}) \\
\bottomrule
\end{tabular}
\vspace{-2mm}
\caption{Abstention performance across different models broken down by \textbf{answerable} and \textbf{unanswerable} questions across various datasets. Baseline abstention rates are shown at the top with a gray background. Abstention rates under (b) No context, (c) Random context, and (d) Noisy context settings are shown below as deltas from the base rate. Colors indicate a \hlpos{positive} or \hlneg{negative} delta. All context perturbations improve model abstention performance in some settings, though this is not uniform over datasets, models, question answerability, or perturbation setting. 
}  
\label{tab:abstention_all}
\end{table*}

%% file: tables/cases_abstention.tex
\begin{table*}[tbhp!]
\setlength\extrarowheight{1pt}
    \small
    \centering
    \scalebox{0.81}
    {
    \begin{tabular}{lp{0.5\linewidth}p{0.5\linewidth}}
    \toprule
    & \textbf{\squad}& \textbf{\qasper} \\
    \midrule
     \textbf{Context}& Computational complexity theory is a branch of the theory of computation in theoretical computer science that focuses on classifying computational problems according to their inherent difficulty, and relating those classes to each other... 
     & Table TABREF35 show the comparisons between tree and sequential based methods. We can see that, if we don't deploy CNN, simple Tree LSTM yields better result than traditional LSTM, but worse than Bidirectional LSTM...
     \\
     \midrule
     \textbf{Prompt}
     & Q: What is a manual application of mathematical steps? & Q: Do they separately evaluate performance of their learned representations (before forwarding them to the CNN layer)?\\
     \midrule
     \textbf{Given context}& Unanswerable & No \textcolor{red}{\ding{55}}\\
    \textbf{No context}& Calculation \textcolor{red}{\ding{55}}& Yes  \textcolor{red}{\ding{55}}\\
     \textbf{Random context}& Unanswerable & Unanswerable \\
      \textbf{Noisy context}& Unanswerable & Unanswerable \textcolor{red}{\ding{55}} \\
     \midrule
     \textbf{Ground truth}& Unanswerable & Yes \\
     \bottomrule
    \end{tabular}}
    \vspace{-2mm}
    \caption{Examples of \chatgptShort predictions under different context perturbations. For \squad, removing context results in the model no longer abstaining, and responding inaccurately. For \qasper, the model answers the question incorrectly in the gold context setting; removing context results in an accurate but incorrect answer (the model should abstain in this case instead).}
    \label{tab:case_study}
\end{table*}

%% file: sections/discussion_new.tex
\vspace{-2mm}
\section{Discussion \& Conclusion}

Our study investigates the impacts of context removal and perturbation on LLM performance on scientific QA. While lack of correct context should result in a model abstaining from answering a question, our results highlight that there are inconsistent patterns of model behavior based on model pretraining paradigms, question types, and the context-dependence of various QA datasets. For example, perturbing given context by replacing with random irrelevant context or adding noisy context would be expected to reduce task performance, but in some cases, the improvements in abstention that result can negate any reductions in task performance and potentially lead to gains (e.g., \vicunaShort and \gpto on \pubmedqa, \llamaShort and \vicunaShort on \qasper). Additionally, we find that abstention varies greatly by question type, with all models in our experiments struggling to abstain on boolean questions.

Future work for enhancing models' abstention ability could investigate the impact of
(i) \textbf{Different prompting strategies}: since model abstention ability is sensitive to constrained prompts (such as boolean instructions) as shown in Section \ref{sec:rq2}, how to select good prompting strategies to produce the best trade-off between task performance and abstention performance remains an interesting problem; (ii) \textbf{Alternate model architectures}: smaller models such as Flan-T5-XL with encoder-decoder architectures performed comparably to larger decoder-only models such as ChatGPT. Further exploration of encoder-decoder architectures or introducing an auxiliary module to foster understanding of context may be helpful; (iii) \textbf{Other context perturbations}: we show initial results that providing noisy context can counter-intuitively improve task performance in some datasets due to interactions with unanswerable questions. How this interacts with retrieval errors or malicious injected context that occur in open-domain QA is a direction that could be explored in future work.


Our results have direct implications for dataset curators and model developers. Benchmark QA datasets with mixtures of unanswerable and answerable questions were designed to facilitate assessment of abstention ability, yet our experiments show conflation and a lack of clear assessment of either abstention ability or performance accuracy. In other words, while unanswerable questions were motivated by the need to measure abstention, when aggregated with other questions during task performance evaluation, this distinction is obscured. Additionally, unanswerable questions in different datasets measure different phenomena, e.g., in \squad, they measure model sensitivity to irrelevant context passages, while in \qasper, they indicate that a question is unanswerable based on the given document, neither of which clearly map to the notion of abstaining under insufficient information. Given the importance of assessing model abstention capabilities, separating task and abstention assessment, coupled with changes in dataset construction, is needed to better align model performance on these tasks with human expectations.

Beyond dataset curation, other variables---such as the task defined by the dataset, the types of questions posed, the architecture of models, instruction tuning techniques, in-context learning, and domain-specific pretraining---can affect a model's ability to effectively abstain and provide accurate answers. In this work, we attempt to disentagle some of these factors, though not all. 
Future studies could explore the extent to which in-domain pretraining and abstention-specific instruction tuning techniques impact model abstention performance.


On the other hand, our results also hold implications for downstream builders/users of interactive systems who rely on LLMs as question-answering tools. 
We show that in some cases (e.g., boolean questions), LLMs exhibit little to no ability to abstain, even for highly context-dependent questions such as those in \qasper. System designers should be cautious of these limitations and when/how to infer model uncertainty and communicate this uncertainty to users. 
For communicating abstention capabilities to users, confidence scores or similar indicators that reflect model certainty, accompanying explanations, or interpretation guidance can empower users to make more informed decisions.





%% file: sections/limitation_new.tex
\vspace{-2mm}
\section*{Limitations} 
This work evaluates LLM abstention in situations with incorrect or noisy context, which mimics retrieval errors in retrieval-augmented systems. The problem of abstaining is also crucial in other settings, such as from the fairness, privacy/copyright, and safety perspectives, which we do not address in this work. 
In the future, we plan to propose a unified framework for abstention evaluation that considers these settings as well.

The interplay between abstention and task performance can only be studied in datasets with unanswerable questions, which is scarce in the landscape of QA datasets. We only conduct experiments on three such datasets, so our results may have difficulty generalizing. Additionally, we note the different ways that unanswerable questions were constructed among these datasets, which imply that our experiments may be measuring different notions of ``unanswerability.'' In addition, although question type diversity is common in real application, there are insufficients numbers of QA datasets with question type diversity. Therefore, our results may not generalize, as several question types are only represented in one or two of our datasets. We also note the sensitivity of LLMs to prompt phrasing, as with constrained and free-form prompts for boolean questions, and emphasize that it may be difficult to acquire consistent results without significant efforts made in prompt engineering.


Due to the budget constraints, we could not conduct experiments with every single model, dataset, or context perturbation method from existing work, though our framework can be applied to other datasets and models. We will make our framework publicly available such that new models and datasets, as well as context perturbations could be added. Other factors affecting abstention such as in-domain pretraining and in-context learning could be incorporated in the future.





%% file: sections/appendix_new.tex
\appendix

\section{Prompting templates}
\label{app:prompt_templates}

\paragraph{Prompts with context} Prompts for other datasets are provided below. For \squad, we do not include a boolean directive as the dataset contains no boolean questions.
~\\ [-3mm]

\noindent Example prompt for \squad:\\ [2mm]
\colorbox{blue!8}{
\begin{minipage}{0.45\textwidth}
\scriptsize
Create a shortest Answer to the Question using the following documents. \textbf{Answer Unanswerable when you are not sure about the answer}. Please only output the exact answer and keep the answer concise. \\ [-2mm]

Context: \{$c$\} \\ [-2mm]

Question: \{$q$\} \\ [-2mm]

Answer:
\end{minipage}} \\

\noindent Example prompt for \pubmedqa:\\ [2mm]
\colorbox{blue!8}{
\begin{minipage}{0.45\textwidth}
\scriptsize
Create an Answer to the Question  using following documents. {\color{blue}{\textbf{Pay attention to answer only ``yes'',``no'' or ``Unanswerable''. }}}\textbf{Answer ``Unanswerable'' when you are not sure about the answer.}\\ [-2mm]

Context: \{$c$\} \\ [-2mm]

Question: \{$q$\} \\ [-2mm]

Answer:
\end{minipage}} \\

\noindent Example prompt for \bioasq:\\[2mm]
\colorbox{blue!8}{
\begin{minipage}{0.45\textwidth}
\scriptsize
Create an Answer to the Question  using following documents. {\color{blue}{\textbf{Pay attention to answer only ``yes'' or ``no''.}}} \textbf{Answer ``Unanswerable'' when you are not sure about the answer.}  \\ [-2mm]

Context: \{$c$\} \\ [-2mm]

Question: \{$q$\} \\ [-2mm]

Answer:
\end{minipage}} \\

\paragraph{Prompts without context} 
~\\ [-3mm]

\noindent The prompting template for no context is different from the original. We remove the instruction expression of ``use the documents'' since no documents are provided. This is the example prompt used for \qasper when context is removed: \\
[2mm]
\colorbox{blue!8}{
\begin{minipage}{0.45\textwidth}
\scriptsize
Create an Answer to the Question. {\color{blue}{\textbf{Answer ``Yes'' or ``No'' for boolean questions.}}} \textbf{Answer ``Unanswerable'' when you are not sure about the answer.} \\ [-2mm]

Context: \{$c$\} \\ [-2mm]

Question: \{$q$\} \\ [-2mm]

Answer:
\end{minipage}} \\

\noindent Correspondingly, we remove this instruction phrase from the prompts for \squad, \pubmedqa, and \bioasq.

\section{Experiment Details}
\label{app:Hyperparameters}
We run all experiments on an A100 GPU with 40GB of memory for \llamaShort, \vicunaShort, \flanShort and \mistral. We set max context length to 3096 and max new tokens to 256. For all models, evaluation batch size is 2. We randomly sample 10\% of \squad test dataset in this work.






\begin{table}[t!]
\centering
\small
\begin{tabular}{lccc}
\toprule
\scriptsize &  \scriptsize \pubmedqa & \scriptsize \bioasq & \scriptsize \qasper \\
\midrule
\multicolumn{4}{c}{(a) Constrained prompt} \\
\midrule
 \llamaShort & (30.3) &  (2.8) & (0.5)  \\
 \vicunaShort & (89.5) & (10.7) & (28.4) \\
 
\flanShort & (0.0) & (0.0) & (0.0) \\
 \chatgptShort & (77.7) & (36.4) & (97.8)  \\
\midrule
\multicolumn{4}{c}{(a) Free-form prompt} \\
\midrule
\llamaShort & (50.3) &  (55.7) & (21.6)  \\
 \vicunaShort & (83.3) & (54.2) & (8.9) \\
\flanShort &  (84.0) & (80.0)  & (92.8)  \\
 \chatgptShort & (48.1) & (29.3) & (43.1) \\
\bottomrule
\end{tabular}
\caption{Abstention performance comparison on boolean questions using constrained prompt and free-form prompt under no context setting. Free-form prompt enables \flanShort to successfully abstain across three datasets. }
\label{tab:freeform}
\end{table}

\begin{table}[t!]
\centering
\small
\begin{tabular}{lcc}
\toprule
\scriptsize &  \scriptsize \pubmedqa & \scriptsize \bioasq  \\
\midrule
\multicolumn{3}{c}{(a) Unanswerable} \\
\midrule
\llamaShort & (30.3) &  (2.8)   \\
 \vicunaShort & (89.5) & (10.7) \\
\flanShort & (0.0) & (0.0)  \\
 \chatgptShort & (77.7) & (36.4)   \\
\midrule
\multicolumn{3}{c}{(a) Maybe } \\
\midrule
\llamaShort & (2.6) &  (0.0)  \\
 \vicunaShort & (9.0) & (0.0) \\
\flanShort & (0.0) & (0.0)   \\
 \chatgptShort & (70.9) & (40.2)   \\
\bottomrule
\end{tabular}
\caption{Abstention performance comparison on boolean questions using different abstention representation (``Unanswerable'' vs ``Maybe'') under no context setting. ``Unanswerable'' enables models to abstain across three datasets.}
\label{tab:maybe}
\end{table}

\begin{table*}[t!]
\setlength\extrarowheight{1pt}
    \small
    \centering
    \scalebox{0.81}
    {
    \begin{tabular}{lp{0.5\linewidth}p{0.5\linewidth}}
    \toprule
    & \textbf{\bioasq }& \textbf{\pubmedqa} \\
    \midrule
     \textbf{Context}& BCL11B mutations in patients affected by a neurodevelopmental disorder with reduced type 2 innate lymphoid cells. Using massively parallel sequencing we identified 13 patients bearing heterozygous germline alterations in BCL11B. Notably, all of them are affected by global developmental delay with speech impairment and intellectual disability; \textbf{however, none displayed overt clinical signs of immune deficiency.} Six frameshift mutations, two nonsense mutations, one missense mutation, and two chromosomal rearrangements resulting in diminished BCL11B expression, arose de novo...
     & ...Postoperative recovery rates in the surgery group at 1 week and 4 weeks were -7.4\% and -1.1\%, respectively. Only 5 cases had showed clinical improvement, and the condition of these 5 patients had worsened again at averaged 7.4 weeks after surgery. Postoperative oral steroid therapy was initiated at an average of 6.4 weeks and the average initial dose was 54.0 mg in the surgery group, while 51.3 mg in the nonsurgery group. The recovery rate of the Japanese Orthopedic Association score, which increased after steroid therapy, \textbf{was better in the nonsurgery group (62.5\%) than in the surgery group (18.6\%) with significant difference (P<0.01)}.
     \\
     \midrule
     \textbf{Prompt}
     & Q: Is there a link between BCL11B haploinsufficiency and syndromic neurodevelopmental delay? & Q: Is decompressive surgery effective for spinal cord sarcoidosis accompanied with compressive cervical myelopathy? \\
     \midrule
     \textbf{Given context}& Yes \textcolor{red}{\ding{55}} & Unanswerable \textcolor{red}{\ding{55}}\\
    \textbf{No context}& Yes \textcolor{red}{\ding{55}}& Yes  \textcolor{red}{\ding{55}}\\
     \textbf{Random context}& No \textcolor{red}{\ding{55}} & Unanswerable \\
      \textbf{Noisy context}& Yes \textcolor{red}{\ding{55}} & Yes \textcolor{red}{\ding{55}}  \\
     \midrule
     \textbf{Ground truth}& No & No \\
     \bottomrule
    \end{tabular}}
    \caption{Examples of \chatgptShort predictions under different context perturbations for \bioasq and \pubmedqa. \chatgptShort tends to hallucinate responses for boolean questions when no or incorrect context is provided. For \bioasq, the model answers the question incorrectly in the gold context setting; removing context results in the model skewing to ``yes'' and random context results in the model answering ``no'', both cases should be ``unanswerable'' ideally. For \pubmedqa, the model answers the question incorrectly in the gold context setting; removing context results in hallucination to ``yes'' (the model should abstain in this case instead).}
    \label{tab:case_study_more}
\end{table*}

\section{Ablation Study} 
\label{app:ablation}

We conduct experiments removing constraints from boolean question prompts. We refer to the resulting prompts as ``free-form'' prompts. The example used for \qasper is given below:\\
[2mm]
\colorbox{blue!8}{
\begin{minipage}{0.45\textwidth}
\scriptsize
Create an Answer  to the Question using following documents. \textbf{Answer ``Unanswerable'' when you are not sure about the answer.} \\ [-2mm]

Context: \{$c$\} \\ [-2mm]

Question: \{$q$\} \\ [-2mm]

Answer:
\end{minipage}} \\

\paragraph{Free-form vs. constrained prompts for boolean questions} The free-form prompt consistently improves the abstention ability of smaller models on boolean questions across datasets as shown in Table~\ref{tab:freeform}. Compared with zero abstention performance (obtained under the contrained prompt), the free-form prompt enables \flanShort to successfully abstain across three datasets; in fact, \flanShort demonstrates the best abstention ability for boolean questions under the free-form prompt setting across datasets. For \llamaShort, the free-form prompt also notably increases its abstention performance.

\paragraph{Different ways to represent abstention} We further ablate the wording used to represent abstention in our prompt (results in Table~\ref{tab:maybe}). Instead of using ``unanswerable'', we also try the term ``maybe'' in the prompt. We find that using ``maybe'' makes models less likely to abstain. This may be due to ``unanswerable'' being used more frequently in training corpora to represent abstention.

\section{Further error analysis} 
\label{app:error_analysis}

Table~\ref{tab:case_study_more} presents example failure cases from \bioasq and \pubmedqa.